# Algorithms for screening of Cervical Cancer: A chronological review

[1] Yasha Singh, [1] Dhruv Srivastava, [2] P.S. Chandranand, [2] Dr. Surinder Singh

[1] JSS Academy of Technical Education, Noida, India
[2] National Institute of Biologicals, Noida, India

## 1. Abstract

There are various algorithms and methodologies used for automated screening of cervical cancer by segmenting and classifying cervical cancer cells into different categories. This study presents a critical review of different research papers published that integrated AI methods in screening cervical cancer via different approaches analyzed in terms of typical metrics like dataset size, drawbacks, accuracy etc. An attempt has been made to furnish the reader with an insight of Machine Learning algorithms like SVM (Support Vector Machines), GLCM (Gray Level Co-occurrence Matrix), k-NN (k-Nearest Neighbours), MARS (Multivariate Adaptive Regression Splines), CNNs (Convolutional Neural Networks), spatial fuzzy clustering algorithms, PNNs (Probabilistic Neural Networks), Genetic Algorithm, RFT (Random Forest Trees), C5.0, CART (Classification and Regression Trees) and Hierarchical clustering algorithm for feature extraction, cell segmentation and classification. This paper also covers the publicly available datasets related to cervical cancer. It presents a holistic review on the computational methods that have evolved over the period of time, in chronological order in detection of malignant cells.

## 2. Introduction

AI is rapidly progressing into many areas of clinical care; nonetheless radiology and other imaging-rich disciplines, such as pathology etc, clearly being its earliest beneficiaries. Machine learning and computer vision is one of the most advanced aspects of AI, and many algorithms are able to surpass the human ability to distinguish extremely fine details in clinical images, with tremendous promise of accuracy for diagnostic, radiology and pathology. Here, we project such methodologies in cervical cancer therapeutics, primarily focussing on detection of malignant cells. Cancer of cervix is the situation in which the cervix, which in normal situation is covered by a thin layers of tissue consisting of normal cells changes into malignant cells which have a tendency to grow and divide more rapidly than usual, thus developing into a tumor. By medical image processing and tissue texture analysis methods, cervical cancer screening at early stage becomes feasible and cancer can be controlled in the the dysplasia itself, before it steps into chronic stage. The intelligent system approach for analysis of malignant cells is cost-effective and found to have an edge over the routine detection methods like Pap smear and Liquid cytology based (LCB) test or Colposcopy or Cervicography etc in terms of accuracy and time.

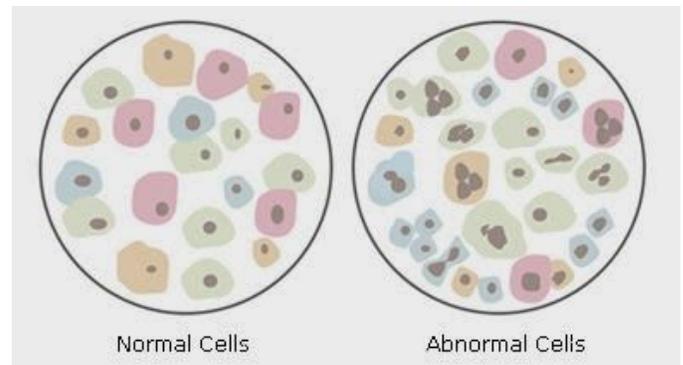

Fig1: A normal and Abnormal (cancerous) cell [22]

## 3. Datasets:

AI applications require maximum volumes of data, which must be well structured i.e. collate, stored securely, normalized, annotated, analyzed, and accessible to end users in a meaningful, intuitive fashion. Cancer detection using intelligent systems involves organised data set, labelled properly to assist accurate training of models. The publically available datasets can be listed as follows -

### 3.1 Herlev dataset

One of the finest and most efficient to process cervical cancer data comes from the Herlev University Hospital, from its department of Pathology in collaboration with the Department of Automation at Technical university Denmark. It contains clear images of pap smear cells divided in all the seven stages of cancer and total 917 in count. Cell images were taken at a magnification of 0.201μm/pixel. Each cell is manually classified into seven classes by clinicians and cyto-technicians as described in Fig 2.

### 3.2. Haceptte data set

This dataset consists of 198 pap test images taken at different magnifications. It contains 82 images taken at 20x magnification, 84 images taken at 40x magnification, 32 images taken at 100x magnification. It was collected from PAP test slides of 18 different patients. Size of each image is 2048x2048. This data is prepared by Department of Computer engineering at Bilkent University and Department of Pathology at Hacettepe University Hospital.

### 3.3. Cervical Cancer: Risk Factors Data set

This data was prepared by 'Hospital Universitario de Caracas' in Caracas, Venezuela and is available at UCI Machine learning repository for download[25]. It contains record of 858 patients considering their habits, demographic information, medical history etc. Since several patients decided not to disclose some of the questions due to privacy concerns, the data has some missing values (if patient smokes, had STDs etc)

## 4. Images of Data Set :

| Normal Cells | | | Abnormal Cells | | | |
|---|---|---|---|---|---|---|
| 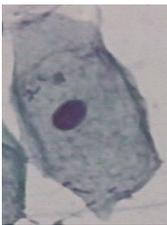 | 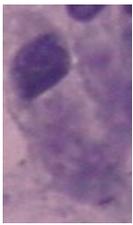 | 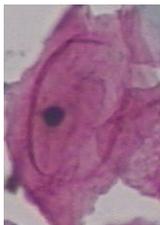 | 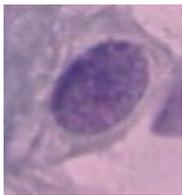 | 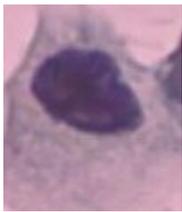 | 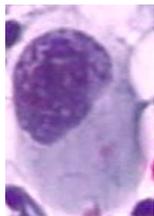 | 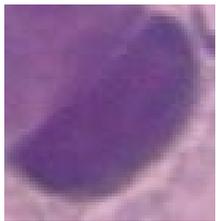 |
| 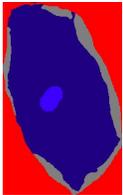 | 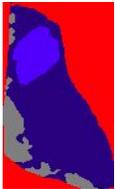 | 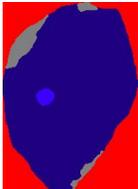 | 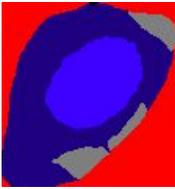 | 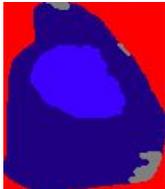 | 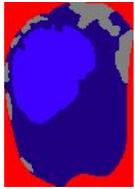 | 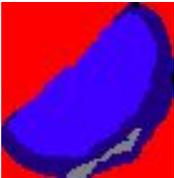 |
| Normal Intermediate | Normal Columnar | Normal Superficial | Moderate Dysplastic | Light dysplastic | Severe Dysplastic | Carcinoma in situ. |

**Fig 2. Herlev Data set images.**

Row first: Images contains the pap smear cell images. Row second: Images contains the nucleus segmented cell images.

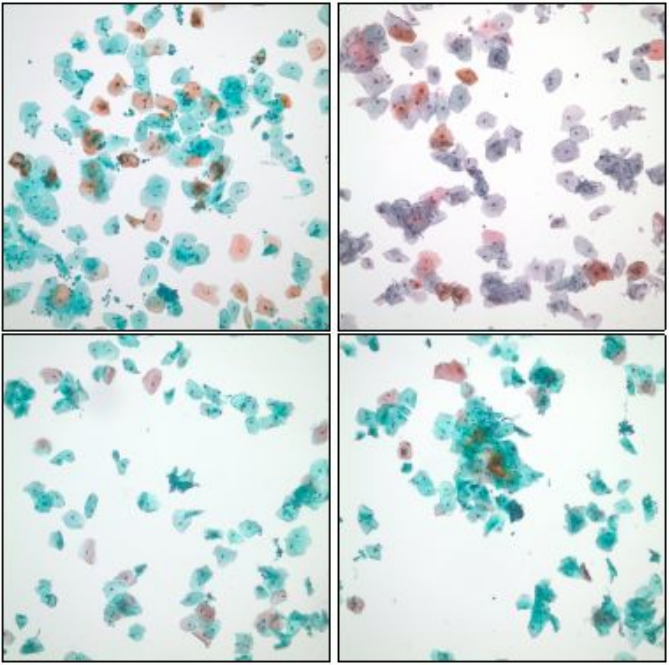

**Fig 3. Images from Haceptte Data set [3]**

## 5. Description of pap smear test and Flow Graph of procedure:

The pap smear test shorthand for Papanicolaou test is essentially a method of screening of cervix and is used to detect cells in the cervix which are malignant or in the process of turning malignant. Cells which have been affected by the HPV (Human Papillomavirus) are examined under microscope in pap smear test. Out of the cells obtained from the cervix, all may be cancerous or only some may or only one cell can be cancerous, in either case it is cervical cancer [12]. In pap-smear test, fixing and staining of the cells is the primary step for the preparation of slide. This prepared slide is then observed under a microscope with a camera and digital images are captured. The digitized images are zoomed while maintaining proper pixel intensity and are then subjected to the process of segmentation and classification using various machine learning and deep learning algorithms. The segmentation step separates the nucleus image from the cytoplasm. The image obtained from this step is called segmented image and it is fed into the classification step as input and it gives decision whether the cell is normal or abnormal.

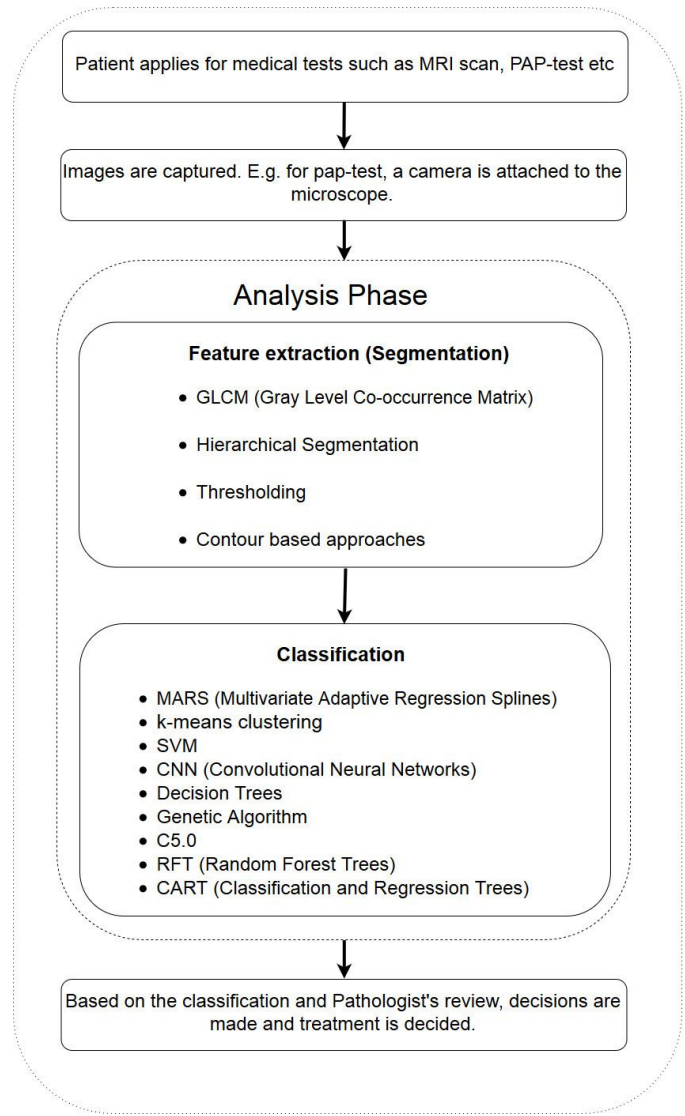

**Fig. 4 Flow graph of cancer detection and classification.**

## 6. Algorithms for screening Cervical Cancer :

<u>1. Support Vector Machines (SVMs)</u>
SVMs are supervised machine learning algorithms majorly used in classification problems. For the case of pap-smear cell classification, this algorithm has proven to yield considerably high accuracy [1]. During training process, it constructs a decision boundary or hyper-planes based on classified input set and divides the points into different classes. In case of high dimensional data, multiple decision boundaries are constructed which separate the data. In real world application where the problem is not linearly separable then such nonlinear problem is mapped to a linearly separable problem.

SVM decision boundary separates the data-points by minimizing the margin value, i.e. the distance between the data-points and the decision boundary, which is based on parameters like kernel, gamma, C, degree etc.

## 2. Hierarchical Clustering

In the research paper by A. Genc¸tav and et al. [3] hierarchical clustering has been used to produce a grouping of cells using features like nucleus area, nucleus brightness, nucleus elongation, cytoplasm area etc. It is a data-mining technique in which a binary tree gets constructed in which each node itself represents another binary tree (cluster) till leaf level. To form a new cluster in the subsequent levels, merge operation is performed on similar clusters. The clusters that are merged are selected based on pairwise distances in the form of a distance matrix [3].

## 3. C5.0

C5.0 algorithm is used for the construction of binary tree. It has evolved from its predecessor C4.5 algorithm. C5.0 creates decision trees from labelled input data (training set). It is generally used for classification problems and selects the best optimised attribute from given information using Max Gain method. [7] C5.0 is much faster, memory efficient compared to C4.5 and offers multithreading for execution. C5.0 algorithm follows approach of greedy algorithm [4].

## 4. Genetic algorithm

In the research paper by S. R.P Singh and et al. Genetic algorithms is used for staging cervical cancer. Genetic algorithm (GA) are metaheuristic, i.e it uses principles of natural selection and follows randomized search space optimization technique. It uses selection, mutation and crossover as its main operators. [5] GA is an iterative algorithm which initiates an initial population (randomly selected) called a generation with each member having set of properties. A value known as fitness is evaluated for each individual which is essentially the value of objective function of that specific problem. Individuals with high fitness values are selected stochastically and properties of every individual are modified leading to a new generation. This leads to newer generation which leads to more optimized solutions.

## 5. Probabilistic Neural Networks

This feed-forward artificial neural network is based on Bayesian Networks which typically consists of four layers- An Input layer, pattern layer, summation layer and an output layer. These computational models are generally used for pattern recognition and classification problems. [6] This model follows Bayes optimal Classification. PNN offer high speed advantage. PNN paradigm has proved to be 200,00 times faster than back-propagation [21]. Use of PNN for cervical cell classification can be observed in [6].

## 6. Convolutional Neural Networks

These feed forward deep neural networks have proved to be a big breakthrough in task of Image Classification. This layered structure mainly uses convolutional layers, pooling layers, normalization layers and dense layers. CNN has proved to be a breakthrough in processing images, video, audio and speech data compared to all types of neural networks [20]. The features extracted by this network are exceptionally effective in object recognition [8]. CNN uses the concept of shared weights hence they are able to process high dimensional data effectively.

## 7. K-means clustering

K-means clustering is a centroid based clustering algorithm in which the formation of clusters is performed based on the proximity of maximum points to each other. It is classified as unsupervised learning algorithm and follows iterative method. In every iteration, the centroid is approximated due to new data-point. This algorithm creates 'K' clusters from 'N' data points such that data points in one cluster have similar features [9]. This algorithm has been used in [9] and [10] for classifying cervical cancer cells into different categories.

## 8. Decision trees

These algorithm provide explicit representation of decision making at every step in the form of hierarchical structures (tree structures) [7]. These structures consist of root node, intermediate nodes and leaf nodes. The root node characterizes the basic condition beginning from which the path ahead is decided. Internal nodes represent further attributes for decision making until a final attribute, leaf node is reached which is the final result of the algorithm. Decision trees can be implemented using basic conditional statements[11].

### 9. RFT (Random Forest Trees)

This algorithm is widely used due to its accuracy and applications in both classification and regression problems. This algorithm builds multiple random decision trees and the resulted ensemble model is used for desired task. For a task, nodes represents an attribute, branches represent an inference and class labels are represented by the leaf nodes. Since data is feeded into different random decision trees, the final class label is calculated by feeding the result of each tree into an non-linear activation function which results in the final class label. This model can handle large datasets and generally does not require hyperparameter tuning. Use of RFT can be witnessed in [9] for classifying stages of cervical cancer.

### 10. CART (Classification and regression Trees)

CART is a recursive algorithm which is used for classification or regression analysis using binary trees [9]. In every binary tree, leaf nodes represent the output variables and other nodes as the input variables and decision steps (cut/split points). Its implementation requires selection of cut/split points and input variables. Cut points are selected which yield the minimum cost as calculated by the problem's cost function. CART algorithm focuses more on problem structure than on data. [9]

### 11. Multivariate Adaptive Regression Splines (MARS)

MARS is a regression model which can efficiently deduce the relationships between a set independent variables and outcome variables (continuously dependent). It is a non-parametric approach which is based on "divide and conquer" procedure, i.e. each input space is categorized by their own regression equation. MARS can handle categorical and continuous data, hence it is more flexible than linear regression. Effectiveness of this algorithm for classifying cervical cancer stages can be observed in [4].

### 12. Gray level co-occurrence matrix GLCM

It is statistical method to analyse images wherein the algorithm relates two pixels in an image by observing their occurence in a specified spatial region. Using these relations it creates a Co-occurrence Matrix (CCM) and then statistical measures are extracted from it [6]. CCM is a matrix wherein each point represents the frequency of the number of times the pixel is observed at that point [10]. GLCM uses this technique to analyse the features. This algorithm is applied in [6] and [10] for texture analysis and feature extraction of MR images.

## 8. Table of Comparison

Among the very first works in cervical cancer detection stands the work by P. Mitra in the year 2000 where staging was done by amalgamation of ID3 and GAs, where GAs were mainly used for refining the architecture. In the work by J. Zhang et al.[1] in 2004 the use of SVMs was more desirable because they had small data set of 40 images containing 149 cells. Back in that time medical imaging data set was meagre and SVMs provide accurate result(clearer decision boundary) on small size batch and also SVM's screening outperformed IG+AVR screening. In 2010, in a work by A. Kale a generic segmentation was defined which permitted correct classification of overlapping or heterogeneously stained cells like in case of Haceptte data and finally classified using support vector machines. In paper by R. Vidya and G. M. Nasira [9], CART algorithm was initially implemented to check the feasibility of the task. On realizing pretty successful results, advanced algorithms RFT and RFT with k-NN were implemented leading up to an accuracy of 96.77%. Staging cervical cancer was not just limited to cell images, MRI scans were also reviewed as observed in the work by K.P. Chandran et al. [6] and M.K. Soumya et al. [10] by texture analysis using GLCM. Lately in 2018, work by L. Zhang et al. [14] showcased the performance of CNN classifier for staging cervical cancer cells over the benchmark Herlev Dataset with high accuracy of 98.6% ± 0.3.

Therefore, in the following table, on the basis of various comparison metrics papers have been scrutinized.

| Title | Dataset | Algorithm | Conclusion | Drawback | Accuracy | Year |
|---|---|---|---|---|---|---|
| Staging of Cervical Cancer with Soft Computing [24] | Numerical data set of 221 patients from CNCI institute. Classified into I-IV stages of cancer with 19, 41, 139, and 19 cases in each. Each set made up 21 Boolean features. | Genetic Algorithms (GA's) and Interactive Dichotomizer 3 (ID3) | The proposed model outperforms basic MLP models in case of complex data. It also reflects the scope of soft-computing in the field of cervical cancer screening. | A simple, low dimensional numerical data was considered for training the proposed model. | 81.5% | 2000 |
| Cervical Cancer Detection Using SVM Based Feature Screening [1] | 149 cell images (41 cancerous and 108 normal) extracted from 149 pap smear test slides. Image size ranges from 93x64 to 300x244 pixels. | Support Vector Machines (SVM) | Improvements in accuracy of pixel-level classification. | Small image database used for generating decision boundary. | 98% | 2004 |
| Segmentation of Cervical Cell Images [2] | Herlev and Haceptte data set | Hierarchical segmentation algorithm | Successfully segmented cells from slide images despite poor staining, poor contrast and overlapping cells. | Nucleus regions could not be extracted from some cells due to noisy texture. | 96% | 2010 |
| Unsupervised Segmentation and Classification of Cervical Cell Images [3] | Hacettepe and Herlev data set. | Cell segmentation involved morphological operations, thresholding followed by hierarchical segmentation algorithm. Classification of cells included hierarchical clustering. | Successfully segmented cells from pap-smear test slides despite poor contrast and accurately classified the cells into different categories. | Issue with segmenting cells with overlapping cytoplasm areas. | -- | 2012 |
| Prediction of Recurrence in Patients with Cervical Cancer Using MARS and Classification [4] | A numerical Dataset prepared by the Chung Shan Medical University Hospital Tumor Registry, with 12 predictor of 168 patients. | C5.0 and MARS (Multivariate Adaptive Regression Splines) | Identified specific risk factors for the recurrent cervical cancer problem using data mining. | Considered a completely numerical dataset. | C5.0- 96% MARS- 86% | 2013 |
| Genetic Algorithms for Staging Cervical Cancer [5] | A set of 239 cervical cancer patient cases that have been obtained from the database of International Gynaecologic Cancer Society (IGCS) | Genetic Algorithm | Genetic algorithms worked efficiently when large amount of information of the cervical cancer patient is provided as input and similar diagnostic case is to be searched from the large amount of data. | Mutation is unguided hence too many generations may be witnessed to get optimal solution. | -- | 2013 |

| Title | Dataset | Method | Description | Limitations | Accuracy | Year |
|---|---|---|---|---|---|---|
| Improving Cervical Cancer Classification On Magnetic Resonance Images using texture analysis and PNNs. [6] | Magnetic Resonance Images (MRI) of cervical cancer | Feature extraction using GLCM (Gray Level Co-occurrence Matrix) and cancer classification using PNNs (Probabilistic Neural Networks). | Features and spatial relations extracted from MRI images proved useful for prediction. | Classification accuracy results are different for different datasets | 92.8571% | 2015 |
| Cervical Cancer stage prediction using Decision Tree approach of Machine Learning [7] | IGCS prepared data of 237 patients with 10 distinct features. | C-5.0 algorithm | Their method classifies the stages of cervical cancer using decision trees. | Data set is case specific. Does not include general tests like pap-test. Low accuracy. | 67.5% | 2016 |
| Pap Smear Image Classification Using Convolutional Neural Network [8] | 2 diagnostic centers prepared an image dataset of 1611 Pap smear cell images and Herlev Dataset. | Convolutional Neural Network | Efficient and accurate identification of dysplasia from Pap smear images using deep learning. | High training time. To increase the accuracy, additional feature extraction methods were involved. | 90-95% | 2016 |
| Prediction of Cervical Cancer using Hybrid Induction Technique: A Solution for Human Hereditary Disease Patterns [9] | 500 records containing 61 variables prepared by NCBI (National Center for Bio-technology Information) consisting of Biopsy numerical values with gene identifiers. | CART (Classification and Regression Tree) algorithm, RFT (Random Forest Tree) algorithm and RFT with K-means learning | Combined RFT with k-means clustering algorithm for cervical cancer prediction to obtain an accuracy of 96.77%. | The data is specially prepared for this method. No general test such as Pap-test is being used. | CART - 83.87%  RFT - 93.54%  RFT with k-mean - 96.77% | 2016 |
| Cervical Cancer Detection and Classification Using Texture Analysis [10] | MRI (Magnetic Resonance Images) of 24 patients divided into axial T1 and T2-weighted images and sagittal T2-weighted images. | Texture analysis (feature extraction) using GLCM and cell classification using SVM | Statistical features and transform features for prediction could be outperformed by using texture features. | Results in different accuracy for different types of MR images (axial and sagittal). | 81-83% | 2016 |
| Machine learning Technique for detection of Cervical Cancer using k-NN and Artificial Neural Network [12] | Herlev dataset | K-Nearest Neighbours (k-NNs) and Artificial Neural Networks (ANNs). | Coupled fuzzy based segmentation techniques with k-NN to classify cells. | Accuracy differs widely for different types of cells. | k-NN-88.04%  ANN- 54% | 2017 |
| Screening of Cervical Cancer by Artificial Intelligence based Analysis of Digitized Papanicolaou-Smear Images [13] | Pap-smear test of about 200 clinical cases identified by multiple healthcare institutions in North India. It consists of 8091 cervical cells images captured from slides using a digital camera attached to a | Multiple backpropagation neural networks | Results of 10 folds cross validation for different algorithms was obtained which demonstrated that AI can be effectively used for mass level screening of cervical cancer. | -- | 95.622% | 2017 |
| | microscope. | | | | | |
| Cervical Cancer Prediction using Data Mining [11] | Numerical data of 50 females with 10 attributes. | Decision Trees. | A decision tree based classifier was designed and tested to identify a patient as low, moderate and high risk of cancer. | Used general attributes not specific to cervical cancer. | -- | 2017 |
| DeepPap: Deep Convolutional Networks for Cervical Cell Classification [14] | HEMLBC and Herlev datasets | Convolutional Neural Networks (CNNs) | Unlike the earlier methods of classification which involved cytoplasm/nucleus segmentation and hand crafted features, CNN automatically extracts deep features from cell images. | 1. Single patch requires 3.5 seconds for evaluation. 2. Nucleus center is per-required for implementation. 3. Few classes of cells were identified incorrectly. | 98.6 ± 0.3 | 2018 |

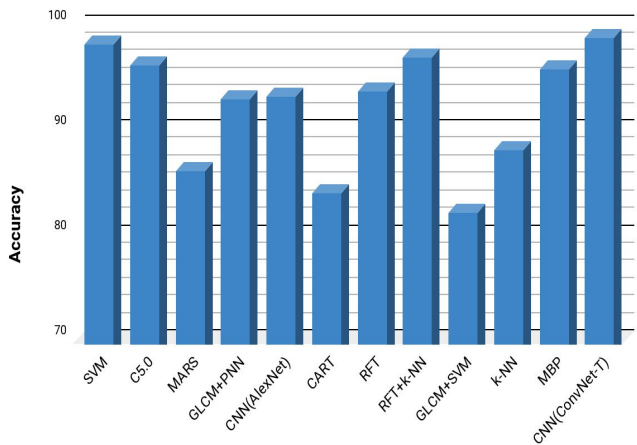

**Fig. 5 Graphical representation of accuracy versus algorithm**

## 8. Conclusion

In conjunction with more accurate diagnostics, AI has the potential to bring down the cost of unwanted interventions for cervical cancer screening. Early detection will promise a greater rate of patients prognosis especially in case of non-invasive cancer. The papers discussed above made use of independent data sources, consequently a base for comparing algorithms on a single scale was hard to define. Multiple algorithms have been applied for segmenting cell cytoplasm, nuclei and other cell components and classifying cells into different categories. In the view of the known stages of cancer, the accuracy of each algorithm, CNN (Convolutional Neural Network) has proved to yield highest accuracy for classifying cervical cell images (PAP smear images) because unlike artificial NN which have fully connected systems, multiple image processing a convoluted task, CNN's accomplish this with greater accuracy because they considerably reduce data-dimensionality, thus the computational overheads. Upon analysis of accuracy of the machine learning algorithms used in the above mentioned papers, it can be inferred that convolutional neural net with ConvNet-T architecture gave maximal accuracy for cervical cell classification. There have been two architectures of convolutional NNs, namely AlexNet and ConvNet-T as addressed in research paper by K. Bora et al. [8] and L. Zhang et al. [14] respectively, **out of which ConvNet-T outperformed all other methods on both datasets including Herlev pap smear and HEMBLC datasets.** *The fact that by making use pre-trained ConvNet-B (made of five conv and three pool layers) in their own model in form of transfer ConvNet-T and additional fully connected layer gave it a lead in fine-tuning over other architectures.* Because ConvNet-T obviated the requirement of data-preprocessing in the form of segmentation. Unlike the approach in paper by K. Bora et al where AlexNet CNN was used for feature extraction followed by least square SVM version and Softmax Regression for classification. Therefore to conclude, Computer Vision methods are potent enough to impact the future trends of cancer diagnostics with high level of expertise.

We implemented transfer learning over "Inception-v2 CNN" mode using benchmark Herlev dataset on a 64bit architecture machine with Linux OS (Ubuntu 14.04) and Intel i7-4710HQ CPU @ 2.50GHz × 8 and a Nvidia GeForce GTX 860M GPU with 4GB memory and 8GB system RAM. We trained the model with over 15,000 iterations which resulted in validation accuracy of

98±0.5% . It took 22 mins and 12s to complete the training process. This verifies the capability of CNN-models for staging cervical cancer cells.

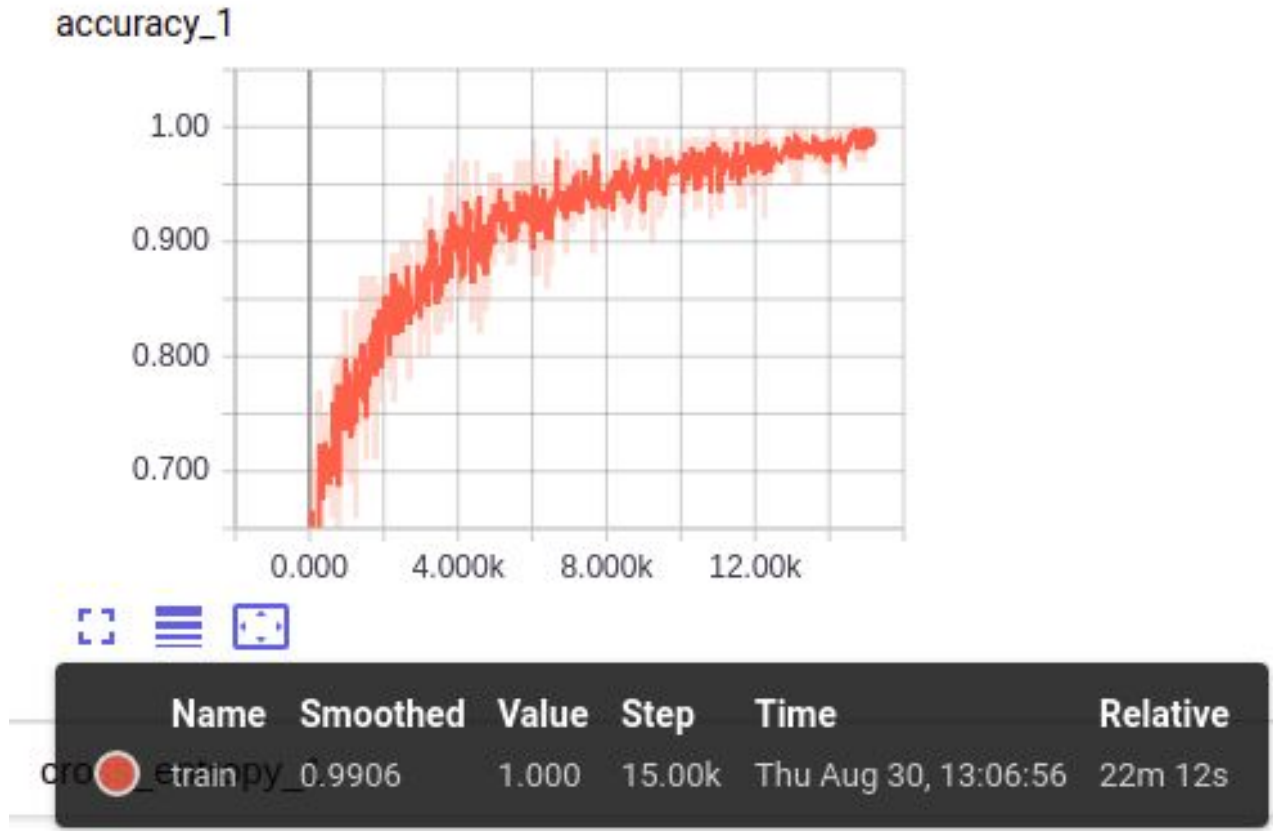